\documentclass[letterpaper, 10 pt, conference]{ieeeconf}  

\IEEEoverridecommandlockouts                             
\usepackage{epsfig} 
\usepackage{mathptmx} 
\usepackage{times} 

\usepackage[utf8]{inputenc}
\usepackage[T1]{fontenc}

\usepackage{graphics} 
\usepackage{amsmath}
\usepackage{multirow}
\usepackage{booktabs}
\usepackage{graphicx}
\usepackage{caption}
\usepackage{pifont}

\usepackage{colortbl}

\usepackage[table,xcdraw]{xcolor}
\usepackage{amssymb}

\newcommand{\YF}[1]{\textcolor{black}{#1}}
\newcommand{\YFF}[1]{\textcolor{black}{#1}}
\definecolor{rblue}{rgb}{0,0.5,1}
\definecolor{awesome}{rgb}{1.0, 0.13, 0.32}
\definecolor{hollywoodcerise}{rgb}{0.96, 0.0, 0.63}
\definecolor{lasallegreen}{rgb}{0.03, 0.47, 0.19}
\definecolor{hanpurple}{rgb}{0.32, 0.09, 0.98}
\definecolor{green(pigment)}{rgb}{0.0, 0.65, 0.31}

\definecolor{mygray}{gray}{.9}
\definecolor{mygreen}{RGB}{93,174,86}

\makeatletter
\newcommand{\thickhline}{%
	\noalign {\ifnum 0=`}\fi \hrule height 1pt
	\futurelet \reserved@a \@xhline
}

\makeatletter
\let\NAT@parse\undefined
\makeatother
\usepackage[pagebackref=false,breaklinks=true,colorlinks,bookmarks=false]{hyperref}
\hypersetup{colorlinks=true,linkcolor={red},citecolor={hanpurple},urlcolor={magenta}}
\makeatletter
\let\NAT@parse\undefined
\makeatother
\usepackage{xcolor}
\definecolor{rblue}{rgb}{0,0.5,1}
\definecolor{hollywoodcerise}{rgb}{0.96, 0.0, 0.63}
\definecolor{lasallegreen}{rgb}{0.03, 0.47, 0.19}
\definecolor{hanpurple}{rgb}{0.32, 0.09, 0.98}
\definecolor{green(pigment)}{rgb}{0.0, 0.65, 0.31}
\usepackage[pagebackref=false,breaklinks=true,colorlinks,bookmarks=false]{hyperref}
\hypersetup{colorlinks,linkcolor={red},citecolor={hanpurple},urlcolor={magenta}} 

\title{\LARGE \bf
Multi-Keypoint Affordance Representation for Functional\\Dexterous Grasping
}

\author{Fan Yang, Dongsheng Luo,  Wenrui Chen$^{*}$, Jiacheng Lin, Junjie Cai,\\Kailun Yang, Zhiyong Li, and Yaonan Wang
\thanks{This work was partially supported by the National Key R\&D Program of China under Grant 2022YFB4701400/2022YFB4701404, the National Natural Science Foundation of China under Grant 62273137, 62473139, No. U21A20518, and No. U23A20341, the Hunan Provincial Research and Development Project under Grant 2025QK3019, the Hunan Science Fund for Distinguished Young Scholars under Grant 2024JJ2027, and the Open Research Project of the State Key Laboratory of Industrial Control Technology, China (Grant No. ICT2025B20).
(\textit{Corresponding author: Wenrui Chen. E-mail: chenwenrui@hnu.edu.cn.)})}
\thanks{F. Yang, D. Luo, J. Cai, W. Chen, K. Yang, and Z. Li are with the School of Artificial Intelligence and Robotics, Hunan University, Changsha 410012, China. (E-mail: ysyf293@hnu.edu.cn.)}
\thanks{J. Lin is with the College of Computer Science and Electronic Engineering, Hunan University, Changsha 410082, China.}
\thanks{W. Chen, K. Yang, Z. Li, and Y. Wang are also with the National Engineering Research Center of Robot Visual Perception and Control Technology, Hunan University, Changsha 410082, China.}}

\begin{document}

\maketitle
\thispagestyle{empty}
\pagestyle{empty}

\begin{abstract}
Functional dexterous grasping requires precise hand-object interaction, going beyond simple gripping. Existing affordance-based methods primarily predict coarse interaction regions and cannot directly constrain the grasping posture, leading to a disconnection between visual perception and manipulation. To address this issue, we propose a multi-keypoint affordance representation for functional dexterous grasping, which directly encodes task-driven grasp configurations by localizing functional contact points. Our method introduces Contact-guided Multi-Keypoint Affordance (CMKA), leveraging human grasping experience images for weak supervision combined with Large Vision Models for fine affordance feature extraction, achieving generalization while avoiding manual keypoint annotations. Additionally, we present a Keypoint-based Grasp matrix Transformation (KGT) method, ensuring spatial consistency between hand keypoints and object contact points, thus providing a direct link between visual perception and dexterous grasping actions. Experiments on public real-world FAH datasets, IsaacGym simulation, and challenging robotic tasks demonstrate that our method significantly improves affordance localization accuracy, grasp consistency, and generalization to unseen tools and tasks, bridging the gap between visual affordance learning and dexterous robotic manipulation. \YFF{The source code and demo videos are publicly available at \url{https://github.com/PopeyePxx/MKA}.}
\end{abstract}

\section{Introduction}
Functional dexterous grasping enables robots to execute complex object manipulations from human instructions. Unlike simple grasping, it requires a dexterous hand to adapt grasp postures and contact different object regions according to the task, involving fine physical interactions between the fingers and the object. For example, in \textit{``Hold Drill''}, all five fingers firmly grasp the drill head, whereas in \textit{``Press Drill''}, the index finger presses the switch while the others stabilize the handle. The core challenge lies in inferring task-relevant contact regions and grasp postures from visual perception.

\begin{figure}[t!]
\centerline{\includegraphics[width=0.5\textwidth]{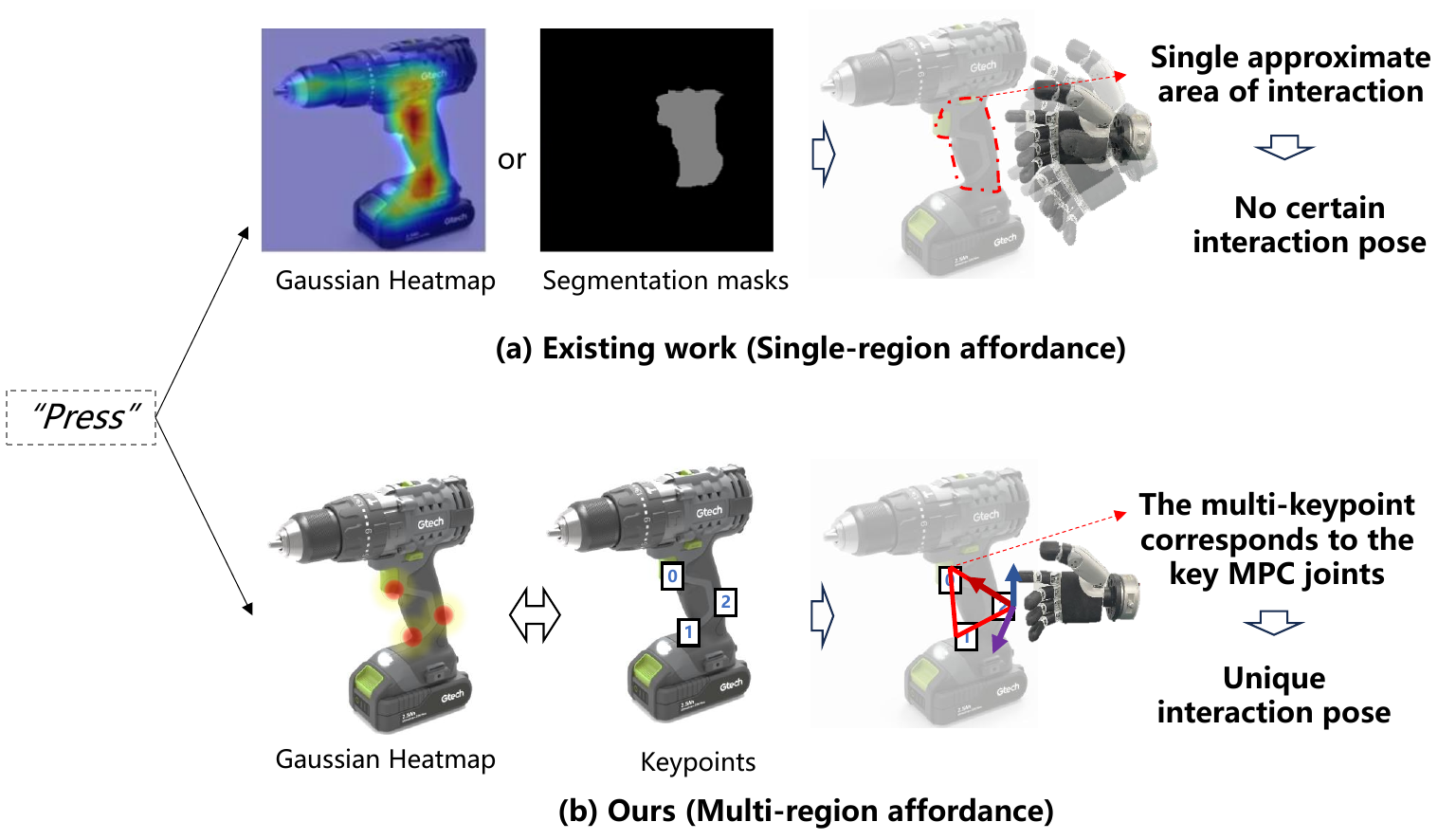}}
\captionsetup{font=small}
\caption{\small Comparison between existing affordance-based grasping methods and our proposed Multi-Keypoint Affordance representation. 
(a) Existing methods identify only a rough interaction region, leading to uncertain interaction poses. 
(b) Our method localizes multiple keypoints corresponding to dexterous hand joints, enabling a precise and constrained grasping posture.}
\label{intro1}
\vskip-2ex
\end{figure}

In the field of vision, affordance-based methods~\cite{gibson1977theory, myers2015affordance,xu2021affordance, luo2022learning,xu2024weakly} have been widely explored to predict potential human interaction regions. Among them, deep-learning-based approaches estimate heatmaps~\cite{luo2022learning,xu2024weakly} or segmentation masks~\cite{myers2015affordance,xu2021affordance} to indicate feasible interaction areas. However, existing methods~\cite{nguyen2024language,li2024learning} can only provide coarse region predictions given an image and a task. A rough affordance map cannot specify the exact interaction posture, leading to uncertainty in the grasping motion and insufficient constraints for functional dexterous grasping, as shown in Fig.~\ref{intro1}(a). Therefore, how to find a novel visual representation that not only identifies task-relevant contact areas but also directly constrains the dexterous grasping posture, ensuring a well-defined interaction between the hand and the object, remains a challenging problem.

Keypoint-based representations offer a potential solution by structuring high-dimensional visual data into a compact and interpretable form. 
Prior work~\cite{koppula2013learning,manuelli2019kpam,qin2020keto,luo2022skp} often decomposes grasping into object and environment keypoints.
For example, KETO~\cite{qin2020keto} defines grasp, functional, and operation points, while SKP~\cite{luo2022skp} specifies five surface keypoints for parallel grasping.
However, these approaches suffer from limited generalization—keypoints are either manually designed for specific tasks, rely heavily on simulation, or require extensive manual annotations, raising data collection costs.

To address these issues, VRB~\cite{bahl2023affordances} learns contact points and motion trajectories from human operation videos, improving generalization and applicability, though it still depends on post-processing and yields indirect visual representations.
Recent advances in Large Vision Models (LVMs) enhance object feature extraction.
For instance, ReKep~\cite{huang2024rekep} automatically detects candidate keypoints via LVMs, filters them with vision-language models, and directly guides robotic operations, improving task generalization and linking vision more directly to action.

Despite successes, the above methods primarily focus on simple two-finger pinch grasps and do not extend to dexterous grasping tasks. In dexterous grasping, keypoints must not only determine the grasping location but also constrain the entire hand configuration, ensuring functional stability, as shown in Fig.~\ref{intro1}(b). Achieving this goal introduces three key challenges: (1) Fine-grained feature extraction: Dexterous grasping involves small, detailed interaction regions between fingers and the object. How can part-level keypoint features be extracted from the object? (2) Data annotation cost: Dexterous grasping requires precise keypoint annotations, which are costly to acquire. How can reliance on manual annotation be reduced? (3) Keypoint correspondence: Establishing a consistent mapping between object keypoints and hand keypoints is essential for stable grasping. How can robust keypoint correspondence be ensured?

To address the challenges, we propose the Multi-KeyPoint Affordance representation for Functional Dexterous Grasping. By localizing multiple keypoints on the object and the hand, a unique dexterous grasping posture with clear constraints is determined. 
First, we introduce the Contact-guided Multi-Keypoint Affordance (CMKA) learning, which leverages LVMs for fine-grained affordance feature extraction. The CMKA supervises Egocentric images using hand-object interaction regions in Exocentric images as contact priors via CAM~\cite{zhang2018adversarial}, guiding keypoint learning towards meaningful functional contact areas and eliminating the need for manual keypoint annotations. 
\YF{Unlike existing methods that only predict contact regions or keypoint locations without providing actionable grasp execution, we propose the Keypoint-based Grasp matrix Transformation (KGT) to bridge perception and control by explicitly computing the relative pose between the hand and the object. Specifically, KGT leverages the geometric relationship among three semantically meaningful keypoints-the wrist, functional finger (index or thumb), and little finger MCP joints-which form a unique triangular structure capturing the relative contact posture.} 
Experiments across $6$ tasks and $18$ tool shapes on the public FAH dataset~\cite{yang2024learning}, achieving an improvement of $45.35\%$ over the state-of-the-art method in the KLD metric. 
In both IsaacGym~\cite{makoviychuk2021isaac} and real robot experiments, we successfully establish the geometric constraint relationship between tool and hand keypoints.

The main contributions of this work are as follows:
\begin{itemize}
    \item A multi-keypoint affordance representation is proposed, which constrains dexterous grasping postures through the geometric relationships of keypoints in the hand-object interaction region, directly establishing a link between vision and dexterous grasping actions.
    \item CMKA, a multi-keypoint affordance localization method based on a weakly-supervised framework, and KGT, a keypoint-based hand-object relative pose transformation method, are introduced, leveraging existing human interaction image data and LVMs for learning, effectively reducing data costs, and enabling functional dexterous grasping.
    \item The proposed algorithm is validated in both simulation and real robot experiments, demonstrating its ability to directly map tasks to grasping actions while exhibiting good generalization across tasks and objects, especially excelling in complex functional grasping scenarios.
\end{itemize}

\section{Related Work}

\begin{figure*}[t!]
\centerline{\includegraphics[width=1\textwidth]{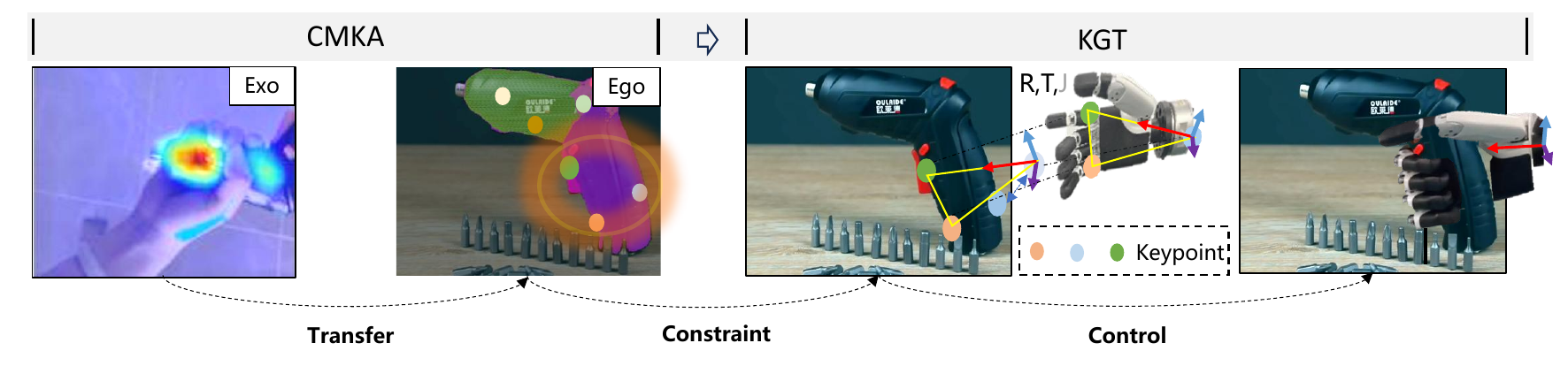}}
\captionsetup{font=small}
\caption{\small The key process of learning and connecting visual perception to functional dexterous grasping actions. The Contact-guided Multi-Keypoint Affordance (CMKA) module learns from human demonstrations in Exo images and transfers this knowledge to Ego views, predicting three keypoints constrained by functional grasp priors. These keypoints are then used by the Keypoint-based Grasp matrix Transformation (KGT) module to compute the hand-object relative pose ((\(R\), \(T\)) to control the grasping task.}
\label{flow}
\vskip-2ex
\end{figure*}

\subsection{Object Representation for Dexterous Grasping}

Grasping and manipulation are fundamental topics in robotics. Traditional methods~\cite{srivastava2014combined, tyree20226, wen2024foundationpose} often rely on six Degrees of Freedom (6DoF) poses to represent objects for parallel gripper tasks. However, these methods are insufficient for dexterous grasping, which requires handling precise contact points on functional regions and complex interactions, going beyond simple object poses that only describe overall position and orientation. Early methods such as rigid body modeling~\cite{rosales2012synthesis, el2015computing} and template matching~\cite{gabellieri2020grasp, kokic2020learning} are task-specific and lack generalization, limiting their applicability to diverse tasks. Recent studies have focused on object structure-based grasp affordance representations, such as ContactDB~\cite{brahmbhatt2019contactdb}, which annotates object-finger contact relationships;
the method in~\cite{zhu2023toward}, which maps contact points to finger regions and intent codes; 
and F2F~\cite{yang2024task}, which uses knowledge graphs to associate functional object parts with functional fingers. While these methods improve task performance, they depend on ideal perception systems that assume precise segmentation or localization of functional regions—an assumption rarely achievable in real-world settings. 
\YFF{In contrast, we propose an object representation specifically designed for dexterous grasping. The object is abstracted as a set of semantically meaningful keypoints on its surface. This structured representation captures functional regions and supports executable grasp synthesis. It provides a bridging link between affordance perception and dexterous grasping control, eliminating the need for idealized perception inputs required by the existing methods~\cite{zhu2023toward, yang2024task}.}

\subsection{Keypoint Representation and Robotic Manipulation}
Keypoint-based methods have been widely applied in computer vision tasks such as face recognition~\cite{mayo20093d, berretti20113d}, human pose estimation~\cite{belagiannis2017recurrent}, and tracking~\cite{chan2017robust}, where keypoints typically serve as low-level features or part-level object descriptors. In robotics, keypoints provide compact representations of the environment and objects. For example, KETO~\cite{qin2020keto} and SKP~\cite{luo2022skp} define different types or fixed numbers of keypoints to describe specific tasks, but these methods often lack generalizability across tasks. 
Recently, ReKep~\cite{huang2024rekep} introduced a more generalizable manipulation framework by leveraging Large Visual Models (LVMs)~\cite{oquab2023dinov2, kirillov2023segment} to extract candidate keypoints and vision-language models to filter task-relevant keypoints for operational guidance. However, ReKep~\cite{huang2024rekep} focuses on simple parallel gripper tasks and requires additional reasoning steps, making it unsuitable for dexterous manipulation. 
Inspired by human hand interactions~\cite{hang2024dexfuncgrasp}, we propose a multi-keypoint representation based on the wrist, functional fingers, and the little finger. This design directly constrains dexterous grasping postures, providing effective and robust solutions for complex manipulation tasks.

\subsection{Visual Affordance and Interaction}
Visual affordance learning explores potential object regions for specific actions and is a key topic in robotic grasping. 
Early fully supervised methods~\cite{nguyen2017object,yang2023grounding} relied on large-scale annotated datasets, which were both expensive and time-consuming to create. To reduce annotation costs, recent research has shifted toward weakly supervised methods, leveraging keypoints~\cite{sawatzky2017weakly,sawatzky2017adaptive} or image-level labels~\cite{luo2022learning,li2023locate,nagarajan2019grounded}. 
In this work, we utilize human interaction images to supervise Ego-view images through contact features, significantly reducing training data costs by leveraging existing resources. Existing affordance methods for robotic manipulation, such as VRB~\cite{bahl2023affordances}, learn contact points and trajectories from human operation videos, whereas Robo-ABC~\cite{ju2024robo} generates hand-object contact datasets to enable zero-shot generalization. Similarly, GAT~\cite{li2024learning} proposes pixel-level affordance learning to capture precise regions. However, those methods often rely on post-processing and additional modules, with coarse affordance regions that lack the fine-grained constraints needed for dexterous grasping.

\begin{figure*}[t!]
\centerline{\includegraphics[width=1\textwidth]{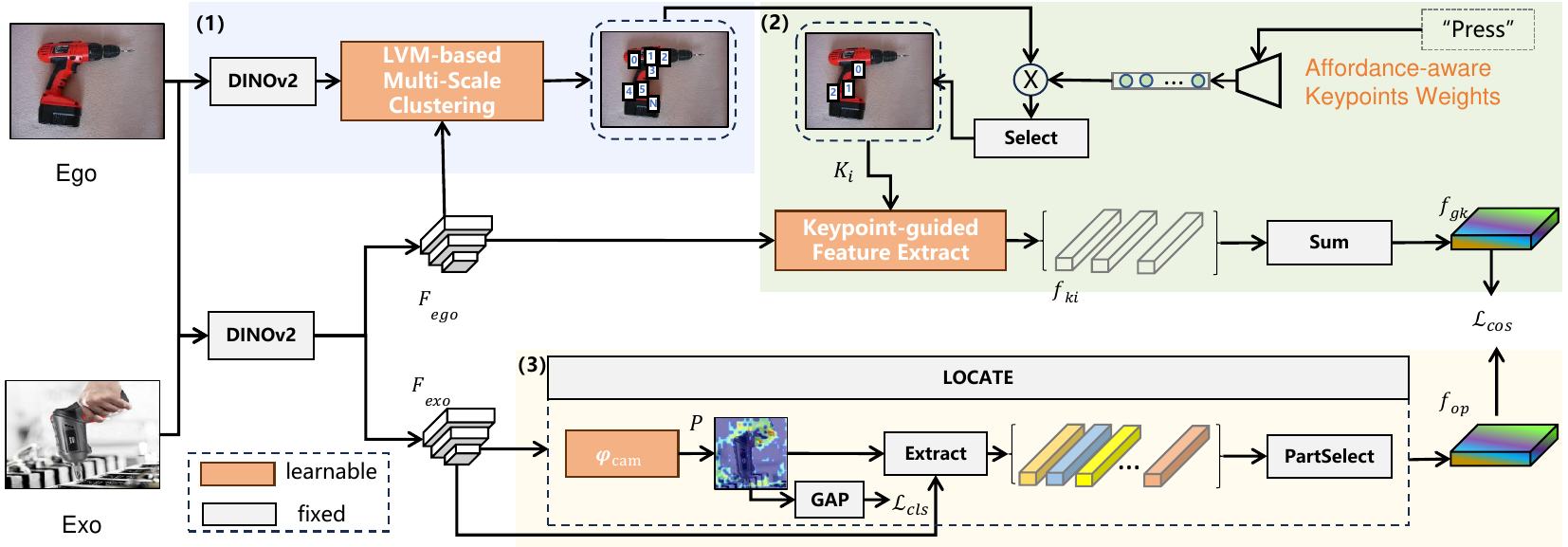}}
\captionsetup{font=small}
\caption{\small \YF{Framework of the proposed CMKA: (1) LVM-based Multi-Scale Clustering for candidate keypoint extraction; 
(2) Affordance-aware keypoint weighting and feature extraction from Ego view; (3) Contact geometry knowledge transfer from Exo to Ego view.}}
\label{pipline}
\vskip-2ex
\end{figure*}

\section{Methodology}
In this work, we propose a complete framework that establishes a direct visual representation for functional dexterous grasping with cross-task and cross-object generalization.  
As illustrated in Fig.~\ref{flow}, the proposed Contact-guided Multi-Keypoint Affordance (CMKA) module learns from human operation experience in exocentric (Exo) images and transfers this knowledge to egocentric (Ego) images, localizing three keypoints constrained by functional dexterous grasping (see Sec.~\ref{cmak}). Using these keypoints, the Keypoint-based Grasp matrix Transformation (KGT) method computes the hand-object relative pose, obtaining the rotation and translation parameters (\(R\), \(T\)) required for grasp execution (see Sec.~\ref{kgt}). During inference, the system requires only an Ego image and the affordance class to predict the three functional keypoints on the object.

\subsection{Contact-guided Multi-KeyPoint Affordances Learning}
\label{cmak}

To identify the keypoint regions on the object surface where the fingers should make contact, robust fine-grained feature extraction is required. 
To achieve this, we first extract multi-level visual features from the Ego and Exo view images using the large vision model DINOv2~\cite{oquab2023dinov2}, which serves as the base visual representation for the following stages. 
As illustrated in the blue-shaded region of Fig.~\ref{pipline}, we then apply LVM-based Multi-Scale Clustering (LMSC) to extract candidate keypoints from different parts of the object surface based on these features (see Sec.~\ref{vmsc}). In the green-shaded region, we design an affordance-aware keypoint weighting and feature extraction mechanism, where weights based on affordance class (\textit{e.g.}, \textit{``Press''}) are assigned to candidate keypoints, guiding both the selection of the most relevant keypoints and their feature extraction from the Ego view (see Sec.~\ref{kfe}). 
In the yellow-shaded region, we leverage the hand-object interaction feature extraction capability of the weakly-supervised LOCATE framework~\cite{li2023locate} for contact geometry knowledge transfer from Exo to Ego view. A cosine similarity loss is applied to supervise the keypoints selection, ensuring that the selected keypoints are concentrated around meaningful hand-object contact regions (see Sec.~\ref{cgkt}).

\subsubsection{\YF{LVM-based Multi-Scale Clustering Module (LMSC)}}\label{vmsc}

\begin{figure}[t!]
\centerline{\includegraphics[width=0.48\textwidth]{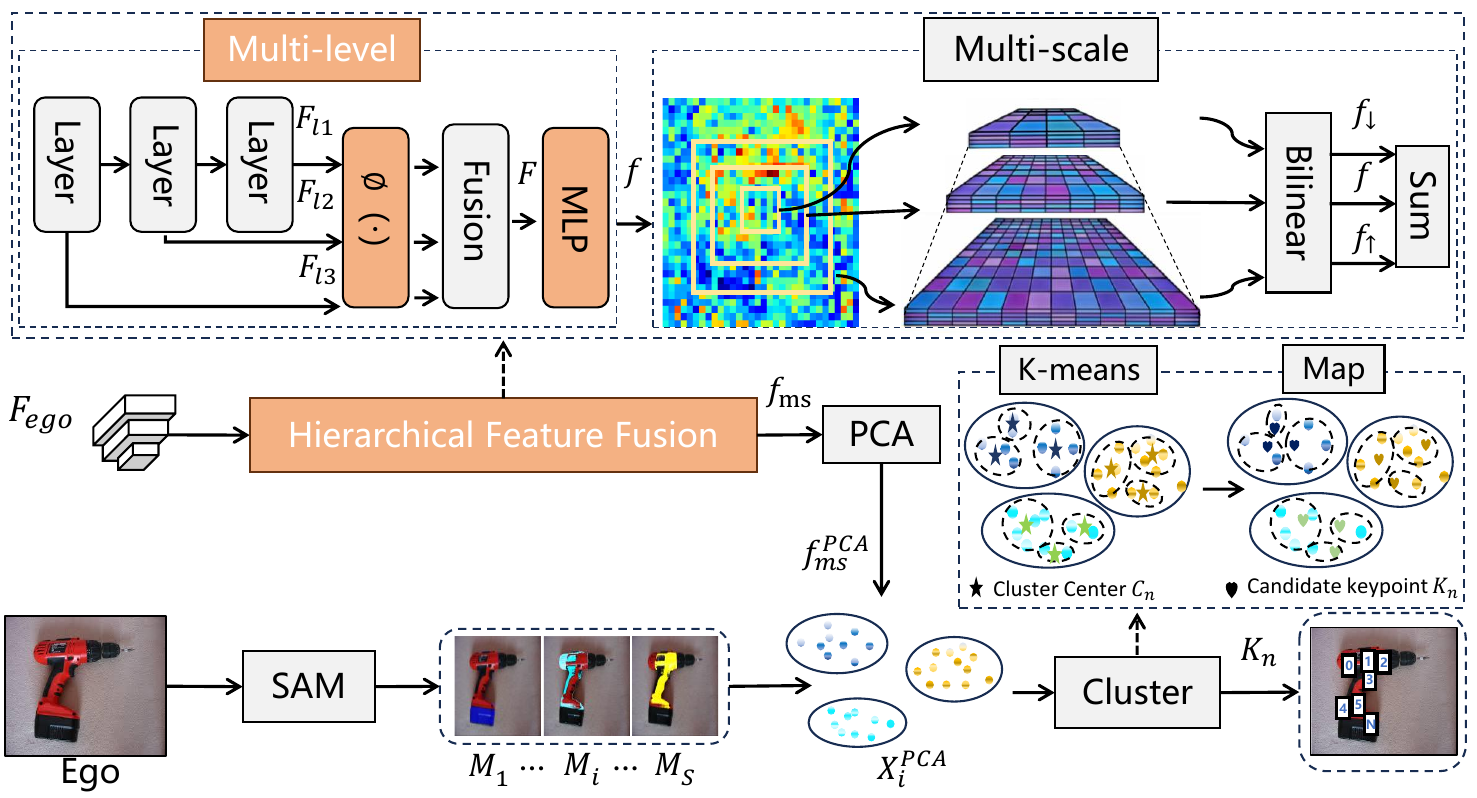}}
\captionsetup{font=small}
\caption{\small \YF{Candidate keypoint generation using the LMSC module.}}
\label{kp}
\vskip-2ex
\end{figure}

\YF{Inspired by ReKep~\cite{huang2024rekep}, we propose the LMSC module to improve keypoints selection in ego-view images, as illustrated in Fig.~\ref{kp}. Given an input image, we first apply the Segment Anything Model (SAM)~\cite{kirillov2023segment} to generate a set of region masks \( \{ M_i \}_{i=1}^{S} \).\\
\indent To improve the perception of fine-grained structural details and contact point features in the object, we then apply a Hierarchical Feature Fusion (HFF) mechanism to integrate multi-level and multi-scale features. Specifically, we extract features \( F_{lm} \) (\( m {=} 1, 2, 3 \)) from the last three layers of DINOv2. We then perform the fusion:
\begin{equation}
F = \sum_{m=1}^{3} \alpha_m \cdot \varnothing(F_{lm}),
\end{equation}
where \( \varnothing(\cdot) \) denotes a linear layer followed by normalization, and \( \alpha_m \) are learnable weights. The fused feature \( F \) is further processed by an MLP to obtain the dense feature \( f \).\\
\indent To capture multi-scale context, we generate bilinearly aligned upsampled features \( f_{\uparrow} \) and downsampled features \( f_{\downarrow} \) versions of \( f \), and sum them with the original to obtain \( f_{\text{ms}} \):
\begin{equation}
f_{\text{ms}} = f + f_{\uparrow} + f_{\downarrow}. 
\end{equation}
\indent Then, we apply Principal Component Analysis (PCA) to \( f_{\text{ms}} \) to obtain a reduced feature map \( f_{\text{ms}}^{\text{PCA}} \), from which we extract region-wise features \( X_i^{\text{PCA}}\).\\
\indent Lastly, we introduce a two-step Cluster module, as illustrated in Fig.~\ref{kp}, to extract candidate keypoints from $X_i^{\text{PCA}}$. Specifically, we first apply K-means clustering on each $X_i^{\text{PCA}}$ to obtain $J$ cluster centers per region, resulting in a total of $N {=} S {\times} J$ cluster centers across all $S$ segmented regions:
\begin{equation}
\{ C_n \}_{n=1}^{N} = \{ C_{ij} \}_{i=1,j=1}^{S,J} = \text{K-means}(\{ X_i^{\text{PCA}} \}_{i=1}^{S}).
\end{equation}
\indent These cluster centers are then mapped back to the pixel space by selecting the nearest feature vector from $X_i^{\text{PCA}}$, yielding the candidate keypoints $\{ K_n \}_{n=1}^{N}$:
\begin{equation}
K_n = \arg\min_{x \in X_i^{\text{PCA}}} \| x - C_n \|^2.
\end{equation}
\indent If clustering is not applicable (\textit{e.g.}, due to insufficient pixels), the geometric centroid of \( M_i \) is used as a fallback.}

\subsubsection{Affordance-aware Keypoint Feature Extraction from Ego View}\label{kfe}
To extract keypoint features from the Ego view image, we define a set of learnable weights \( W {\in} \mathbb{R}^{T \times N} \), where \( T \) represents the number of affordance classes and \( N \) is the number of candidate keypoints. These weights are multiplied with the candidate keypoints \( K_{n} \) to select the final three keypoints \( K_i \) (where \( i {=} 1, 2, 3 \)) for feature extraction from the corresponding regions in the Ego view image.

For the selected keypoints \( K_i \), we define a circular region centered at each keypoint with a radius \( r \) and extract features from these regions, denoted as \( F_{ki} \).

\YF{To align the features from the Ego and Exo views in a unified feature space, we project the features of the selected three keypoints and sum them to obtain the final keypoint feature \( f_{gk} \), which encapsulates the contact geometry information from the Ego view:
\begin{equation}
f_{gk} = \sum_{i=1}^{3} \text{proj}(F_{ki}),
\end{equation}
where \( \text{proj} (\cdot) \) denotes a linear projection layer.}

\subsubsection{Contact Geometry Knowledge Transfer}\label{cgkt}
To supervise egocentric keypoints selection, we define an affordance-specific activation function $\varphi_{\text{cam}}$~\cite{zhang2018adversarial}, which consists of a feed-forward layer, two convolutional layers, and a $1 {\times} 1$ class-aware convolution. Given the exocentric feature map $F_{exo}$, $\varphi_{\text{cam}}$ produces the affordance localization map $P {\in} \mathbb{R}^{C \times H \times W}$, where $C$ is the number of affordance classes.  
We apply global average pooling on $P$ to obtain class-wise affordance scores, which are optimized using a cross-entropy loss $L_{\text{cls}}$.\\
\indent In addition, we adopt the Extract and PartSelect modules from LOCATE~\cite{li2023locate} to obtain the object-part prototype feature $f_{\text{op}}$, by filtering part embeddings through clustering and metric, resulting in representative contact features.

Next, we calculate the cosine similarity loss \( L_{\text{cos}} \) between the Exo prototype features \( f_{\text{op}} \) and the global Ego keypoint features \( f_{\text{gk}} \):
\begin{equation}
L_{\text{cos}} = 1 - \frac{f_{\text{op}} \cdot f_{\text{gk}}}{\| f_{\text{op}} \| \| f_{\text{gk}} \|}.
\end{equation}

The final loss is the combination of the classification loss and the cosine similarity loss, ensuring that the contact geometry knowledge is accurately transferred between the two views:
\begin{equation}
L = L_{\text{cls}} + L_{\text{cos}}.
\end{equation}

\subsection{Keypoint-based Grasp Matrix Transformation}\label{kgt}
\begin{figure}[!t]
\centerline{\includegraphics[width=0.48\textwidth]{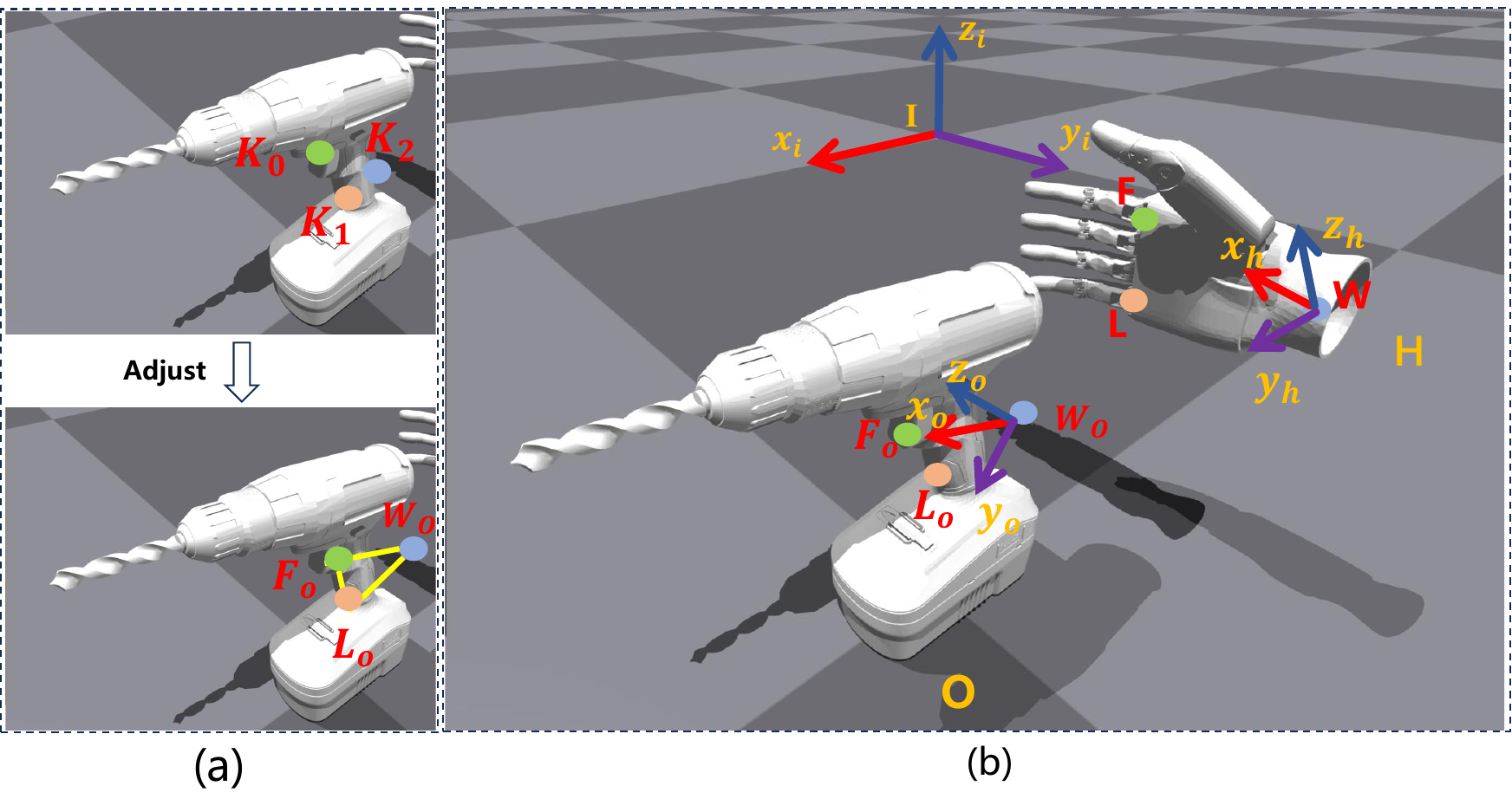}}
\captionsetup{font=small}
\caption{\small Illustration of KGT method in IsaacGym~\cite{makoviychuk2021isaac}, showing the keypoints on the object and the hand (functional finger, little finger, and wrist) and their role in coordinate frame construction.}
\label{hto}
\vskip-2ex
\end{figure}

After obtaining the three keypoints \( K_i \) on the object, we apply the KGT to obtain the relative pose transformation matrix \((R, T)\) between the dexterous hand and the tool. Specifically, as shown in the Fig. \ref{hto} (a), we take \( K_0 \) as the reference point, determine the direction from \( K_0 \) to \( K_1 \), and form a plane using \( K_0 \), \( K_1 \), and \( K_2 \). Based on the hand model (yellow triangle), we adjust the keypoints, resulting in the corrected contact points positions in the world coordinate system \( F_o \), \( L_o \), and \( W_o \). 

Then, as shown in the Fig.~\ref{hto}(b), we define the object coordinate system \( O \) with \( W_o \) as the origin, the x-axis as \( \mathbf{x}_o {=} \frac{\overrightarrow{W_oF_o}}{\|\overrightarrow{W_oF_o}\|} \), the z-axis as \( \mathbf{z}_o {=} \frac{\overrightarrow{W_oF_o} \times \overrightarrow{W_oL_o} }{\|\overrightarrow{W_oF_o} \times \overrightarrow{W_oL_o}  \|} \), and the y-axis as \( \mathbf{y}_o {=} \mathbf{z}_o {\times} \mathbf{x}_o \), leading to the object rotation matrix in the world coordinate system:
\begin{equation}
R_O^I = [\mathbf{x}_o, \mathbf{y}_o, \mathbf{z}_o].
\end{equation}
At the same time, we obtain the transformation matrix between the world coordinate system \( I \) and the object coordinate system \( O \):
\begin{equation}
T_O^I =
\begin{bmatrix}
 R_O^I &   W_o  \\
 0 & 1
\end{bmatrix}.
\end{equation}
Similarly, the hand coordinate system \( H \) is defined with \( W \) as the origin, the x-axis as \( \mathbf{x}_h {=} \frac{\overrightarrow{WF}}{\|\overrightarrow{WF}\|} \), the z-axis as \( \mathbf{z}_h {=} \frac{\overrightarrow{WF} \times \overrightarrow{WL} }{\|\overrightarrow{WF} \times \overrightarrow{WL}  \|} \), and the y-axis as \( \mathbf{y}_h {=} \mathbf{z}_h {\times} \mathbf{x}_h \), where \( F \), \( L \), and \( W \) represent the  keypoints positions on the hand in the world coordinate system, yielding the hand rotation matrix:
\begin{equation}
R_H^I = [\mathbf{x}_h, \mathbf{y}_h, \mathbf{z}_h].
\end{equation}
The relative rotation matrix between the hand and the object is then computed as:
\begin{equation}
R = (R_O^I)^{-1} R_H^I,
\end{equation}
while the translation vector is given by:
\begin{equation}
T = (T_O^I)^{-1}( W-W_o).
\end{equation}

\section{Experiments}
\begin{figure*}[t!]
\centerline{\includegraphics[width=0.9\textwidth]{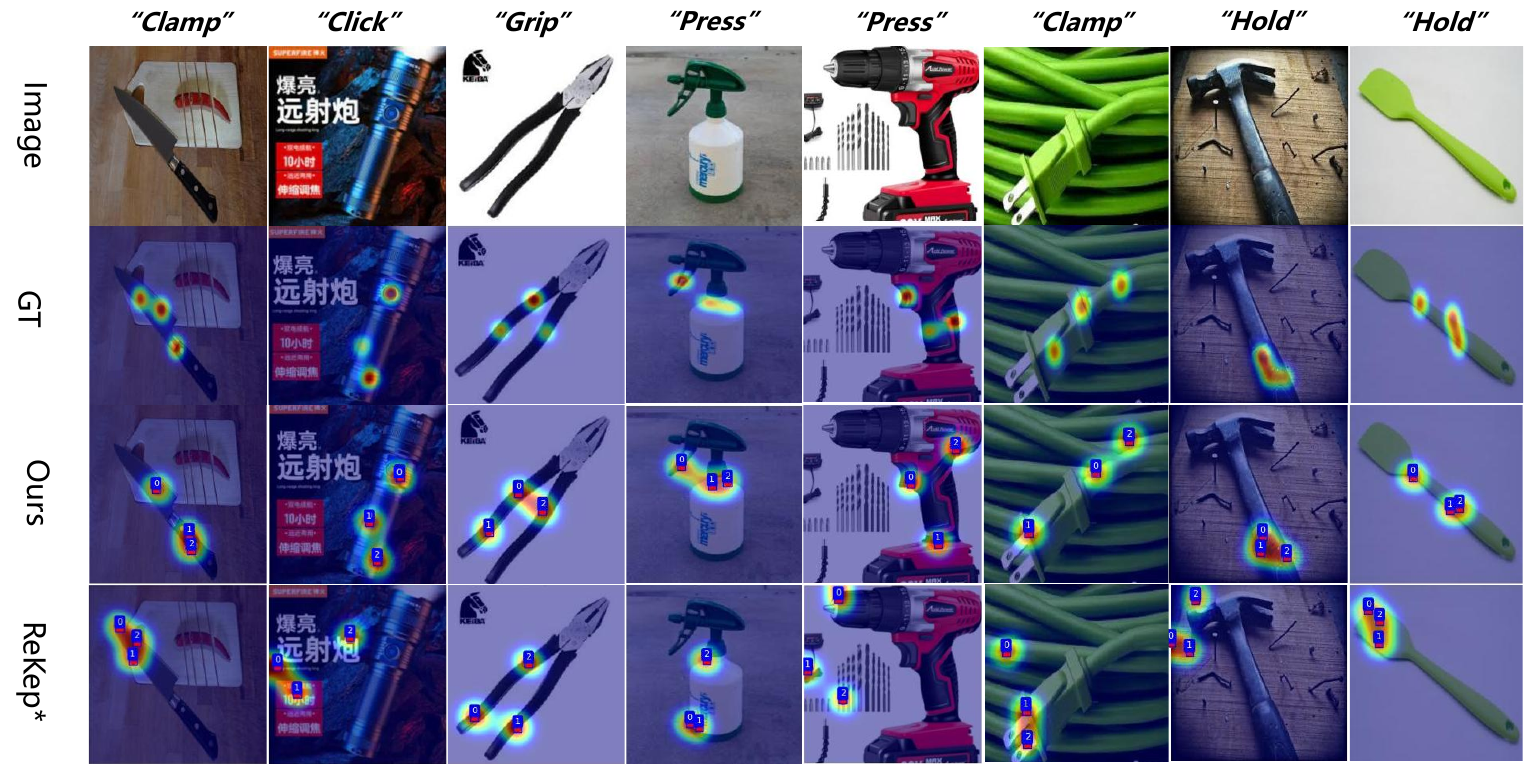}}
\captionsetup{font=small}
\caption{\small Qualitative comparison between our approach and the state-of-the-art multi-keypoint affordance grounding method (ReKep*~\cite{huang2024rekep}) on the FAH test set~\cite{yang2024learning}.
Our proposed method predicts keypoints that are more concentrated in the contact areas and captures the geometric information of the grasping posture. }
\label{vs1}
\vskip-2ex
\end{figure*}

\subsection{Experiment Setup}

\textbf{Datasets:} 
The public challenging FAH benchmark~\cite{yang2024learning} is a large-scale affordance-annotated dataset specifically designed for hand-object interactions. 
It contains $6$ functional grasp affordances \YF{(such as \textit{``Press''}, \textit{``Click''}, \textit{``Hold''}, \textit{etc.})}. and $18$ tools, with $5,858$ images spanning both Exo and Ego views. The dataset provides image-level affordance labels for weakly supervised learning and annotations for coarse dexterous grasp gestures targeting specific ``\textit{Task Tool}'' pairs. 
However, its test set only includes heatmap annotations for functional finger contact regions. To address this limitation, we annotate two additional contact points (little finger and wrist projection areas). 
Specifically, polygons with up to five points are constructed around finger keypoints within a radius of $5mm$, and Gaussian kernels are applied at each point to generate dense annotations. During training, point annotations are added to the object regions in each Ego-view image to distinguish foreground and background during segmentation with SAM~\cite{kirillov2023segment}. 

\textbf{Implementation Details:} Experiments are conducted on an NVIDIA RTX A6000 GPU. The model is trained using the SGD optimizer with a learning rate of $0.01$ over $15$ iterations. Images are resized to a resolution of \( 448{\times}448 \). 

\YF{\textbf{Metrics:} For affordance grounding, we adopt Kullback-Leibler Divergence (KLD), Similarity (SIM), and Normalized Scanpath Saliency (NSS) following prior work~\cite{luo2022learning, li2023locate}.\\
\indent For functional grasping, we measure the functional grasp success rate (FGS) as defined in~\cite{zhu2023toward}: a grasp is successful if the hand holds the object stably for at least ten seconds and correctly performs the intended action on the tool's functional area.\\
\indent In addition, we introduce the 2D-to-Physical Contact Consistency (TPC) metric to evaluate whether the predicted 2D keypoints, when projected into 3D, fall within the functional contact region. TPC is defined as the ratio of the number of hits $n$ to the three selected keypoints.}

\subsection{Results of Functional Affordance Grounding}
In this section, we present qualitative and quantitative results to demonstrate the effectiveness of our method on the FAH test set~\cite{yang2024learning}. 
As weakly supervised methods for multi-region affordance localization are scarce in the state of the art, we use ReKep*~\cite{huang2024rekep}, a keypoint prediction method, as our baseline.

\textbf{Quantitative Results.} 
As shown in Tab.~\ref{pre_table}, our method significantly outperforms ReKep*~\cite{huang2024rekep} across multiple metrics. Specifically, it improves KLD by $45.35\%$, increases SIM by $54.19\%$, and improves NSS by $101.63\%$. 
These improvements stem from ReKep*'s lack of adaptation to dexterous grasping. 
While ReKep*~\cite{huang2024rekep} originally relies on manually selected keypoints, its modified version ReKep*~\cite{huang2024rekep} randomly generates three keypoints without explicit modeling of functional contact regions. In contrast, our method employs a learnable weighting mechanism to generate keypoints specifically for dexterous grasping, ensuring their alignment with functional contact regions. 

\begin{table}[!t]
\centering
\captionsetup{font=small}
\caption{\small Comparison to state-of-the-art method on the FAH test set~\cite{yang2024learning}. 
The \textbf{best} results are highlighted in bold. (↑/↓ means higher/lower is better).}
\label{pre_table}
\begin{tabular}{@{}lccc@{}}
\toprule
\textbf{Model  }                           & \textbf{KLD} ($\downarrow$) & \textbf{SIM }($\uparrow$) & \textbf{NSS} ($\uparrow$) \\ \midrule
ReKep*~\cite{huang2024rekep} & 9.213              & 0.203      & 0.429   \\
Ours & \textbf{5.035}  & \textbf{0.313}    & \textbf{0.865}\\ \bottomrule
\end{tabular}
\vskip-2ex
\end{table}

\begin{figure}
    \centering
\includegraphics[width=0.48\textwidth]{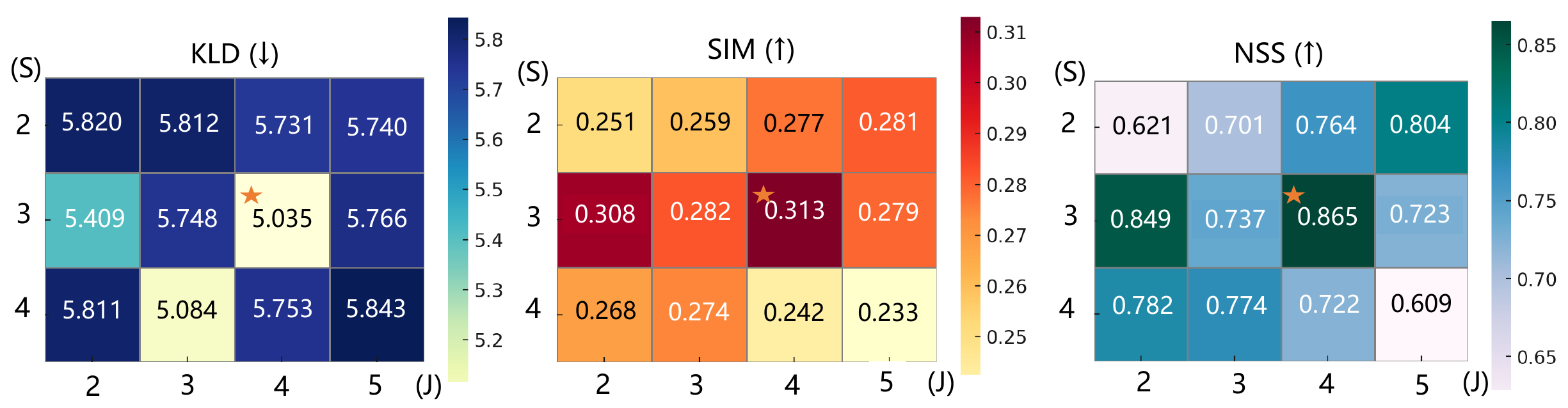}
\caption{\YFF{Performance heatmaps for combinations of region number $S$ and cluster number $J$. The best results are consistently achieved at $S{=}3$, $J{=}4$ (marked by \ding{72}).}}
\label{sjn}
\vskip-2ex
\end{figure}

\textbf{Hyperparameter Sensitivity Analysis.} 
To evaluate the impact of the total number of candidate keypoints $N{=}S{\times}J$ on model performance, we systematically test different combinations of region numbers $S {\in} \{2,3,4\}$ and cluster centers per region $J {\in} \{2,3,4,5\}$, using KLD, SIM, and NSS as evaluation metrics (see Fig.~\ref{sjn}). Experimental results show that the configuration $S{=}3$, $J{=}4$ (\textit{i.e.}, $N{=}12$) achieves the best performance across all three metrics.

In general, moderately increasing the number of candidate keypoints helps improve spatial coverage and semantic diversity, enabling the model to better perceive task-relevant regions. However, an excessive number of keypoints may introduce redundancy and noise, thereby hindering the learning of discriminative weights. Notably, under the condition $S{=}3$, the performance of $J{=}3$ is not only lower than that of $J{=}4$, but also slightly worse than $J{=}2$. This indicates that under weak supervision, the model benefits more from structurally coherent and well-distributed candidate arrangements rather than simply increasing the number of keypoints.

\textbf{Ablation Study.} 
The object priors provided by SAM~\cite{kirillov2023segment} are crucial for constraining keypoint proposals to objects in the scene rather than the background. Thus, we focus on analyzing the critical visual feature extraction network in CMKA. As shown in Tab.~\ref{ablation}, DINOv2~\cite{oquab2023dinov2}, as the backbone network combined with our designed Hierarchical Feature Fusion (HFF) module, achieves the best performance. In the backbone network, DINOv2 generates clearer features compared to DINO-ViT~\cite{DBLP:journals/corr/abs-2112-05814}, better distinguishing fine-grained object regions and leading to improved performance. Furthermore, compared to replacing HFF with a simple fully connected network, HFF, with its multi-layer and multi-scale feature mapping, demonstrates superior potential.

\begin{table}[t!]
    \centering
    \captionsetup{font=small}
    \caption{\small Ablation study on different feature extractors. FFL: feed-forward layer. HFF: hierarchical feature fusion.}
    \begin{tabular}{@{}l@{\hskip 5pt}l@{\hskip 5pt}l@{\hskip 5pt}l@{\hskip 5pt}|@{\hskip 5pt}c@{\hskip 5pt}c@{\hskip 5pt}c@{}}
        \toprule
        \textbf{DINOv2} & \textbf{DINO-ViT}  & \textbf{HFF} & \textbf{FFL} & \textbf{KLD} ($\downarrow$) & \textbf{SIM} ($\uparrow$) &\textbf{ NSS} ($\uparrow$) \\
        \midrule
        \textbf{\checkmark} & &\textbf{\checkmark}  & &\textbf{5.035} & \textbf{0.313} & \textbf{0.865} \\
      \textbf{\checkmark} & & &\textbf{\checkmark}& 5.517 & 0.302 & 0.67 \\
        &\textbf{\checkmark} &\textbf{\checkmark} & & 5.807 & 0.267 & 0.77\\
        &\textbf{\checkmark} & & \textbf{\checkmark}  & 6.075 & 0.253 & 0.65 \\
        \bottomrule
    \end{tabular}
    \label{ablation}
    \vskip-2ex
\end{table}

\textbf{Qualitative Analysis.} 
As shown in Fig.~\ref{vs1}, we compare visibility grounding results from Ground Truth (GT), our method, and the baseline ReKep*~\cite{huang2024rekep}. Our method accurately localizes keypoints within the hand-object contact region while preserving the spatial relationships among the functional finger, little finger, and wrist, ensuring a meaningful distribution for dexterous grasping. For example, in \textit{``Click Flashlight''}, keypoints correctly align with the thumb and little finger, while in \textit{``Press Drill''}, they align with the index finger, little finger, and wrist. In contrast, ReKep*~\cite{huang2024rekep}, which relies on manual post-processing, fails to constrain keypoints within the contact region and lacks spatial consistency, resulting in scattered and less reliable affordance localization.

\subsection{Evaluation of Keypoint-Based Grasp Transformation}
\begin{figure}[t!]
\centerline{\includegraphics[width=0.5\textwidth]{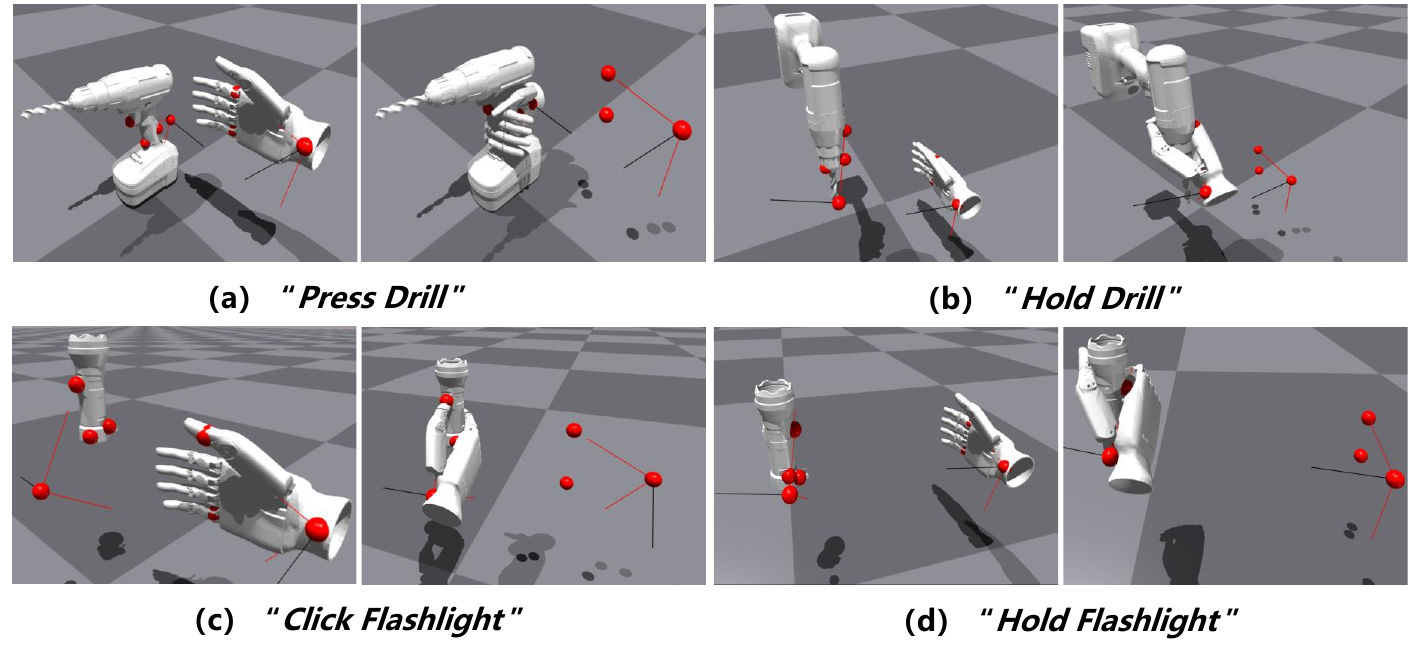}}
\captionsetup{font=small}
\caption{\small Visualization of initial and final hand-object states in IsaacGym~\cite{makoviychuk2021isaac} for different ``\textit{Task Tool}'' combinations. The red spheres represent the three keypoints used for grasp transformation.}
\label{issac}
\vskip-1ex
\end{figure}

To validate the effectiveness of the keypoint-based dexterous grasp transformation method KGT, we conduct experiments on four ``\textit{Task Tool}'' combinations: \textit{``Press Drill''}, \textit{``Hold Drill''}, \textit{``Click Flashlight''}, and \textit{``Hold Flashlight''}. 
As shown in Fig.~\ref{issac}, we visualized the initial and final hand-object states in the simulation environment IsaacGym~\cite{makoviychuk2021isaac}. The results demonstrate that our method accurately computes the grasp transformation matrix, enabling precise hand-object interaction across different task-tool combinations with varying initial hand-object relative postures. For functional interaction tasks, such as \textit{``Press Drill''} and \textit{``Click Flashlight''}, the method ensures correct contact between the functional fingers and the target components. For general grasping tasks, such as \textit{``Hold''}, our method achieves a natural grasp, ensuring a reasonable hand posture.

\subsection{Performance in Real-World Scenarios}

\begin{figure}[!t]
  \centering
  \small
  \includegraphics[width=0.48\textwidth]{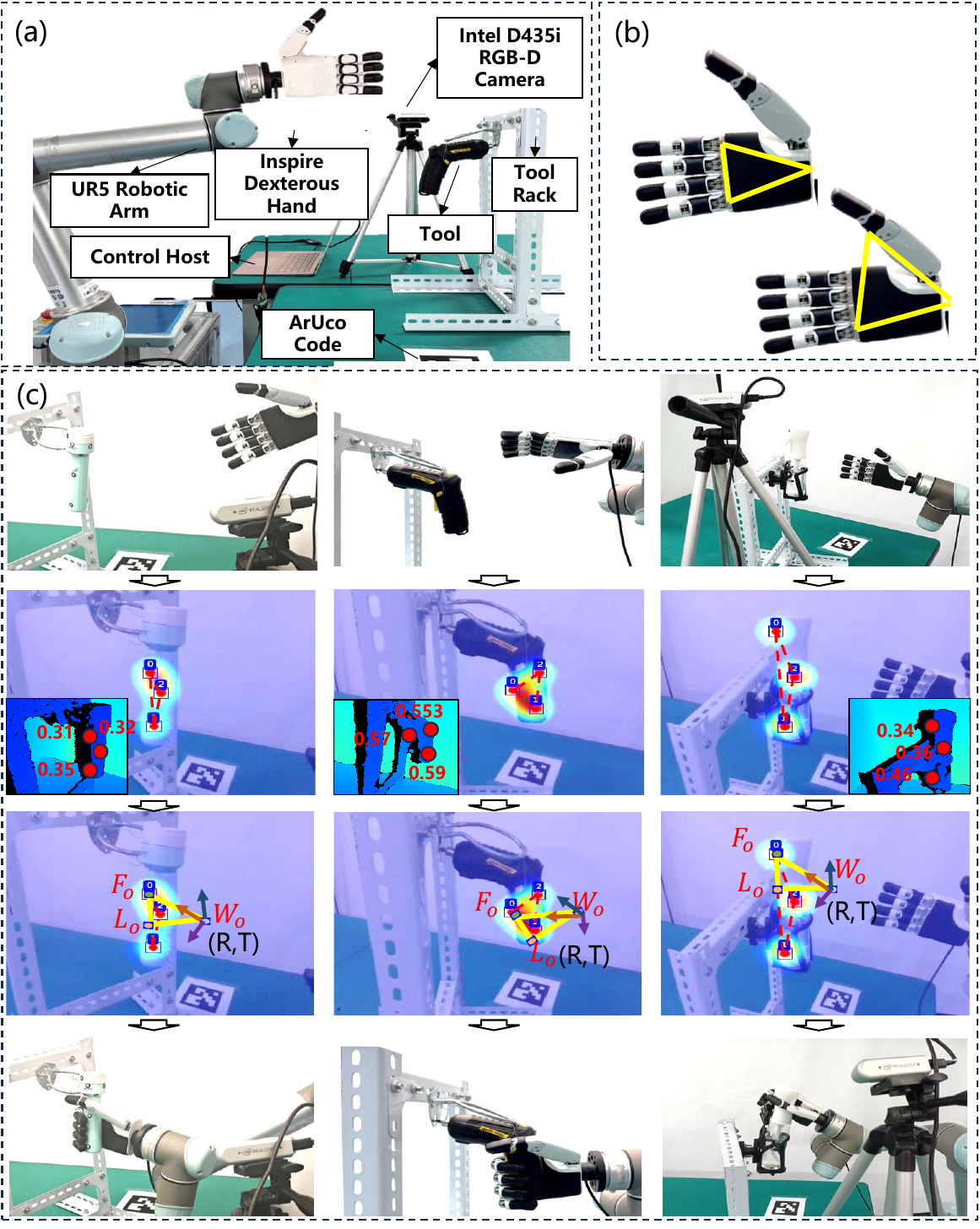}
  \captionsetup{font=small}
  \caption{\small \YF{Experiments with three typical ``\textit{Task Tool}'' in real-world scenarios. (a) Hardware setup. (b) Calibration standards illustrating the geometric relationship between keypoints when the functional finger is the index finger (up) and the thumb (down). (c) Functional grasping steps across the 1st to 4th rows: Initial, Localization, Adjustment, and Control.}}
 \label{fig:teaser}
\end{figure}

As shown in Fig.~\ref{fig:teaser} (a), the real-world platform consists of an Inspire hand, a UR5 industrial robotic arm, an Intel RealSense camera, a tool rack, and a control computer. To address real-world uncertainties, we introduce keypoint relative position calibration annotations based on the Inspire model during the grasping process, as shown in Fig.~\ref{fig:teaser}(b).

In the experiments, we evaluated common daily tools unseen during training, including three \textit{``Task-Tool''} pairs with strict functional grasping requirements: \textit{``Click Flashlight''}, \textit{``Press Drill''}, and \textit{``Press Spraybottle''} (Fig.~\ref{fig:teaser}(c)). The process includes four steps: (1) CMKA localizes three affordance keypoints and estimates their planar relationship; (2) depth values at the 2D locations are retrieved from the RGB-D depth map to obtain 3D coordinates \YF{(see the bottom of the second row)}; (3) keypoints are adjusted on the hand palm based on calibration (see the yellow triangles in the third row); (4) KGT computes the wrist grasp pose $W (R, T)$, combined with a coarse grasp angle $J$ from FAH~\cite{yang2024learning}, to execute the grasp. Results show that our method effectively bridges perception and execution, enabling precise finger-object alignment in complex tasks like \textit{``Press''} and \textit{``Click''}.

\begin{table}[t!]
    \centering
    \captionsetup{font=small}
    \caption{\small \YF{Results of FGS / TPC (\%) on three representative \textit{``Task-Tool''} in real-world scenarios.}}
    \begin{tabular}{@{}lccc@{}}
        \toprule
        \textbf{Method} & \textbf{\textit{Click Flashlight}} & \textbf{\textit{Press Drill}} & \textbf{\textit{Press Spraybottle}} \\
        \midrule
        GAAF-Dex~\cite{yang2024learning} & 60 / -- & 0 / -- & 0 / -- \\
        \textbf{Ours} & \textbf{80 / 66.7} & \textbf{60 / 100} & \textbf{40 / 66.7} \\
        \bottomrule
    \end{tabular}
    \label{fs}
    \vskip-2ex
\end{table}

Furthermore, due to the lack of direct methods combining perception and dexterous grasping, we compared our method with the state-of-the-art GAAF-Dex~\cite{yang2024learning} by the functional grasp success rate (FGS).  As shown in Tab.~\ref{fs}, cross the three complex tasks, our method achieves an average FGS of 60\%, compared to 20\% for GAAF-Dex, corresponding to an absolute improvement of 40 percentage points. GAAF-Dex relies on the similarity between initial and target grasp orientations, succeeding only in cases like \textit{``Click Flashlight''}, but failing in others due to the lack of rotation adaptation. In contrast, our method handles arbitrary initial poses.

Tab.~\ref{fs} reports the 2D-to-Physical contact consistency (TPC) results. Our method achieves $100\%$ on \textit{``Press Drill''} and $66.7\%$ on both \textit{``Click Flashlight''} and \textit{``Press Spraybottle''}, showing reliable contact alignment across tasks, though performance varies due to challenges like depth ambiguity and transparency. While high TPC does not ensure high FGS, the results demonstrate good generalization of our method and provide a solid basis for further improvement with feedback and online adjustment.

\section{Conclusion and Future Work}
This work proposes a keypoint-based affordance representation for functional dexterous grasping. By leveraging human experience data for weak supervision and integrating the CMKA module with large visual models,  the proposed method achieves precise multi-point contact localization with pose constraints for functional grasping, effectively reducing annotation costs and improving generalization. The KGT method enables the mapping of dexterous hand postures to object keypoints, ensuring a direct connection between perception and action. Experimental results demonstrate that our method outperforms existing approaches in both localization accuracy and functional grasp success rate. Real-world experiments further show that, while 2D vision alone provides reliable perception, it may be insufficient to ensure robust grasp execution under real-world uncertainties. 

In the future, we aim to utilize multimodal information to enhance the accuracy and stability of multi-point 3D localization in real-world scenarios.

\bibliographystyle{IEEEtran}   
\bibliography{ref} 
\end{document}